\documentclass[letterpaper]{article} 
\usepackage{paper}  
\usepackage{times}  
\usepackage{helvet}  
\usepackage{courier}  
\usepackage[hyphens]{url}  
\usepackage{graphicx} 
\usepackage{natbib}  
\usepackage{caption} 
\usepackage{multirow}
\usepackage{enumitem}
\usepackage{algorithm}
\usepackage{algorithmic}
\usepackage{bm}
\usepackage{multirow}
\usepackage{color}
\usepackage{colortbl}
\usepackage{cuted}
\usepackage{booktabs}
\usepackage{amsmath}
\usepackage{amssymb}
\usepackage{algorithm}
\usepackage{algorithmic}

\usepackage{newfloat}
\usepackage{listings}
\DeclareCaptionStyle{ruled}{labelfont=normalfont,labelsep=colon,strut=off} 
\floatstyle{ruled}
\newfloat{listing}{tb}{lst}{}
\floatname{listing}{Listing}
\title{Trash to Treasure: 
	Low-Light Object Detection via Decomposition-and-Aggregation}
\author{
	Xiaohan Cui, Long Ma, Tengyu Ma, Jinyuan Liu, Xin Fan, Risheng Liu
}
\affiliations{
	Dalian University of Technology
}

\usepackage{bibentry}

\begin{document}

\maketitle

\begin{strip}
	\vspace{-2cm}
	\footnotesize
	\centering
	\begin{tabular}{c@{\extracolsep{0.2em}}c}
		\includegraphics[height=0.252\textwidth]{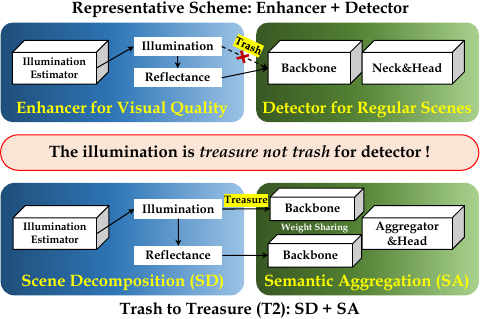} 
		&\includegraphics[height=0.252\textwidth]{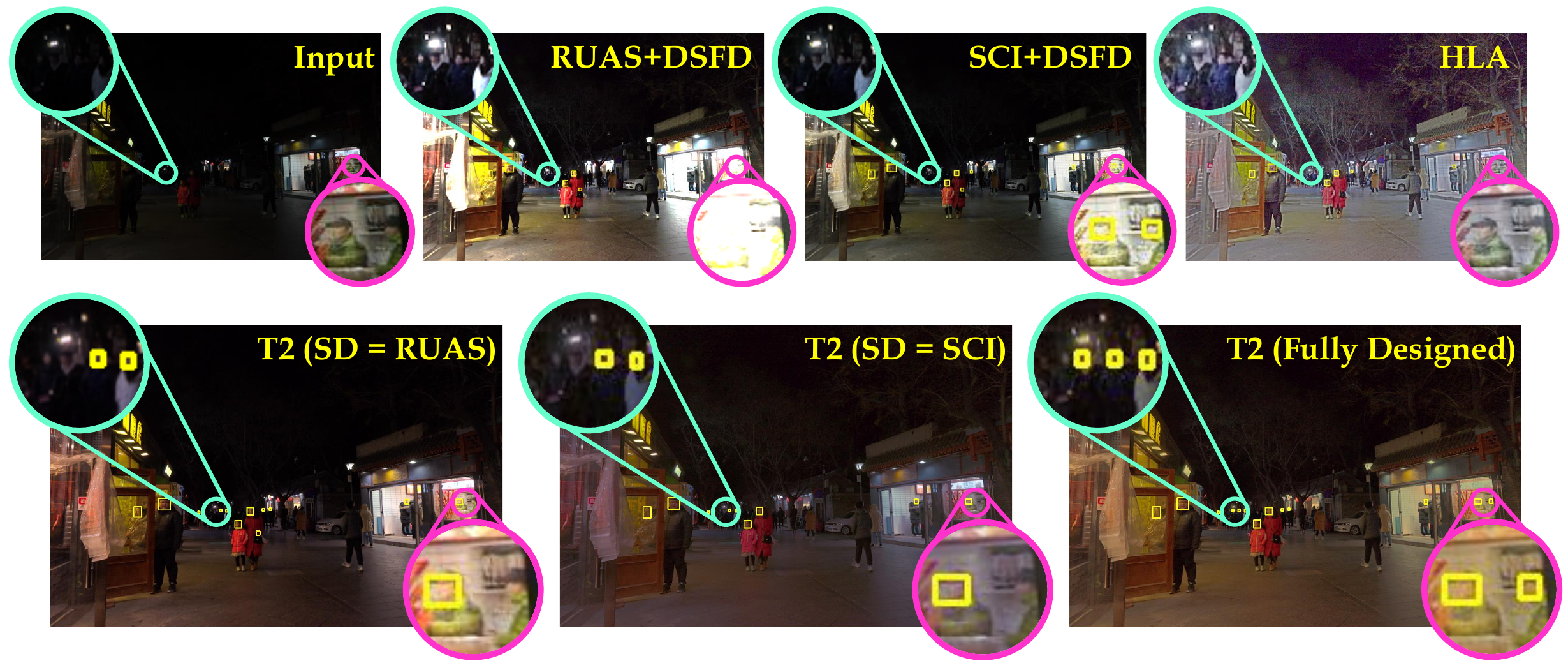}\\
	\end{tabular}
	\vspace{-0.3cm}
	\captionof{figure}{Comparing algorithmic pipelines (\textit{left}) and visual results (\textit{right}). Different from the representative scheme (i.e., enhancer + detector) of low-light detection, we treat the illumination which is viewed as the trash as the treasure, to establish our decomposition-and-aggregation process. It needs to be emphasized that our defined scene decomposition can be instantiated as any illumination-based enhancers. The results in the top row of the right part are two methods based on the representative scheme (enhancer is defined as RUAS and SCI, and the detector is set as DSFD), and a recent method (HLA), respectively. The results in the bottom row are generated by our T2 with different settings of scene decomposition. We can observe that our T2 significantly upgrades the detection ability against the representative scheme. Additionally, our method brightens images (without any loss constraint for enhancement) to realize the enhancement exactly wanted for the detector. }
	\vspace{-0.3cm}
	\label{fig:firstfigure}
\end{strip}

\begin{abstract}
	Object detection in low-light scenarios has attracted much attention in the past few years. A mainstream and representative scheme introduces enhancers as the pre-processing for regular detectors. However, because of the disparity in task objectives between the enhancer and detector, this paradigm cannot shine at its best ability. In this work, we try to \textit{arouse the potential of enhancer + detector}. Different from existing works, we extend the illumination-based enhancers (our newly designed or existing) as a scene decomposition module, whose removed illumination is exploited as the auxiliary in the detector for extracting detection-friendly features. A semantic aggregation module is further established for integrating multi-scale scene-related semantic information in the context space. Actually, our built scheme successfully transforms the ``\textit{trash}'' (i.e., the ignored illumination in the detector) into the ``\textit{treasure}'' for the detector. Plenty of experiments are conducted to reveal our superiority against other state-of-the-art methods. The code will be public if it is accepted. 
\end{abstract}

\section{Introduction}

Object detection is a representative, familiar vision task both in industrial and academic communities. Along with the development of deep learning techniques, object detection in the regular environment has achieved prominent achievements. However, in specific settings, such as salient object detection~\cite{piao2019depth, piao2020a2dele, zhang2020select} and low-light object detection, there remains immense potential for research and optimization. Especially, images taken in low light conditions (e.g., nighttime) usually contain complex degradation to heavily limit the performance of regular detectors~\cite{he2016deep, jin2021bridging, tang2020blockmix, liu2021swin, tang2022learning,  liu2021learning, liu2020real}. This degradation can be due to factors such as noise, low contrast, and insufficient illumination, which can reduce the visibility of objects and make them hard to detect. In the following, we will review existing works and introduce our contributions.  

\subsection{Related Works}
In the past few decades, various schemes have emerged for handling low-light object detection, which can be roughly divided into two categories. The mainstream one is to cascade enhancer and detector, the other is to specifically-design the low-light detector.  


For the enhancer-introduced schemes, the well-known UG2+ Challenge competition\footnote{\url{http://www.ug2challenge.org/}} is a landmark event to drive research for low-light face detection. The cascade of enhancement and detection is the most common solution in the two consecutive championship schemes, e.g., the CAS-Newcastle team adopted this cascaded scheme~\cite{yuan2019ug}, which illustrates the significance of researching the cascaded pattern. In addition, the latest enhancement schemes (e.g., ~\cite{guo2020zero, liu2022learning, ma2022toward, xue2022best, ma2023bilevel}), and the work~\cite{lv2021attention} are no longer satisfied with the improvement of visual quality and try to apply the well-designed enhancer to the detection task to verify the effectiveness of the algorithm. For example, the improvement of detection performance on the Dark Face\cite{yang2020advancing, liu2021benchmarking} dataset is a crucial experiment to demonstrate enhanced performance. However, these works often bring limited performance improvement, and it is even better to directly detect low light images. An intuitive explanation is that these enhancements are designed for human-friendly visual quality\cite{lore2017llnet, wei2018deep, chen2018learning, li2018lightennet}, and are difficult to apply to machine vision tasks focused on high-level semantic information, e.g., object detection. Certain methods produce black edges, retain dark noisy areas, or enhance contrast to improve the overall visual quality, but these enhancements can have a negative impact on the performance of object detection. 


The other type of scheme makes an effort to start from the perspective of designing detectors in an end-to-end manner. Through a bidirectional low-level adaptation and multi-task high-level adaptation scheme, HLA~\cite{wang2021hla, wang2022unsupervised} proposed a joint high-low adaptation framework. This method converted a face detector trained under normal light into a face detector under low light. While domain adaptation methods can mitigate the issue of having a limited number of labeled datasets, their performance improvements were often modest. The method proposed in REG~\cite{liang2021recurrent} is to seamlessly couple a cycle exposure generation module and a multiple exposure detection module, which improved the detection effects by effectively suppressing uneven illumination and noise problems. Moreover, the work~\cite{cui2021multitask} did not apply the enhancement directly to the low-light image, but used the traditional camera signal processing method to transform the normal-light image into a low-light image, and utilized a predictive transform decoder to predict the parameters involved in the illumination transformation to complete the self-supervised training. \cite{liu2022image} proposed a method called IA-YOLO that improves object detection performance in adverse weather conditions. Moreover, clustering helps in extracting better features~\cite{liu2012fixed, wu2019essential}. However, these specifically-designed methods lack the full exploration of scene information, resulting in limited performance improvement.

\subsection{Our Contributions}
To overcome the above drawbacks and arouse the potential of enhancer + detector, this work establishes a new detector with decomposition-and-aggregation by fully exploiting the illumination that is viewed as the trash in the previous schemes. As shown in the left part of Figure~\ref{fig:firstfigure}, we compare the algorithmic pipeline with the representative cascaded scheme. Our proposed \textbf{T}rash to \textbf{T}reasure (\textbf{T2}) is acquired around the fact that ``the illumination is a treasure not trash for the detector'', to make full use of the complete scene information. Benefiting from the flexibility of our designed scene decomposition, existing illumination-based enhancers can be plugged into T2. The right part in Figure~\ref{fig:firstfigure} demonstrates the visual results among different state-of-the-art methods. It can be easily observed that T2 not only acquires the best detection accuracy but also significantly ameliorate the detection accuracy for existing methods. Our main contributions can be summarized as
\begin{itemize}
	\item By deeply analyzing the latent relationship between enhancer and detector, we conclude two key challenges for the cascaded pattern. The one challenge is how to realize detection-oriented enhancement, instead of pure human eye-friendly visual quality. The other is how to reduce information discrepancy between enhanced output and regular data as much as possible. 
	
	\item By rethinking illumination-based enhancers, we construct a scene decomposition module to acquire two scene-related components for characterizing the low-light scenarios. It supports learning detection-oriented enhancement without introducing additional training constraints related to visual quality.
	
	\item To effectively utilize the decomposed components, we design a semantic aggregation module that is composed of a weight-sharing extractor and a multi-scale aggregator. It fully exploits the scene-related content in the feature space to reduce the information discrepancy between regular and low-light data. 
	
	\item Extensive experimental evaluations are performed to verify our superiority in detection accuracy against other state-of-the-art methods. A series of algorithmic analyses indicate that our T2 successfully realizes the intended target, i.e., detection-oriented enhancement. 
	
\end{itemize}

\begin{figure*}[t]
	\centering
	\begin{tabular}{c}
		\includegraphics[width=0.985\textwidth]{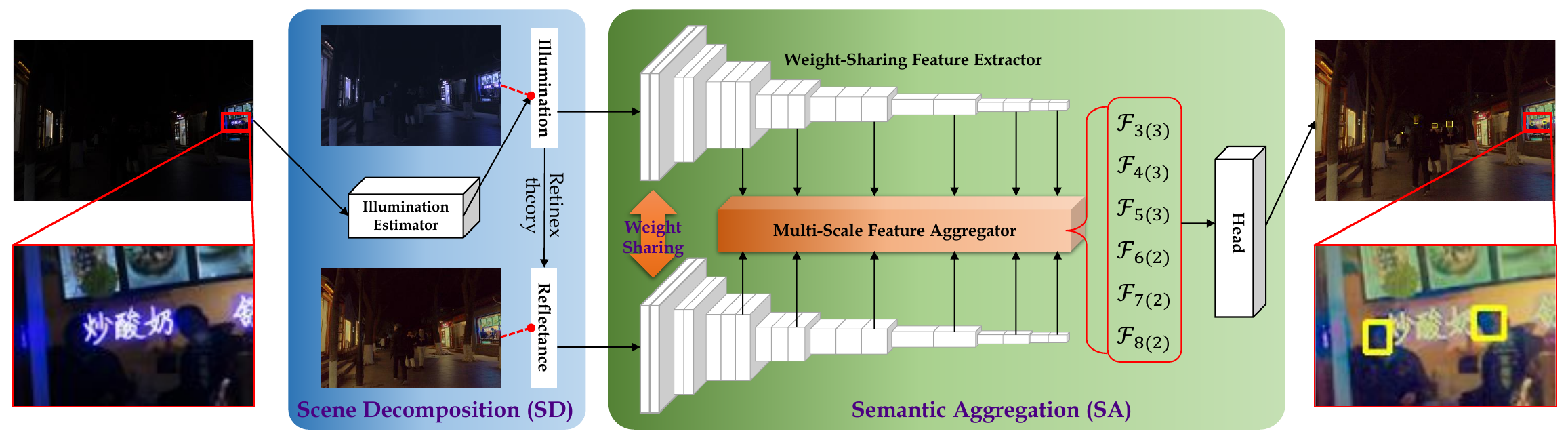}
	\end{tabular}
	\vspace{-0.4cm}
	\caption{The overall architecture of our proposed method. Our method mainly contains two parts, Scene Decomposition (SA) for generating illumination and reflectance from low-light observation based on Retinex theory, and Scene Aggregation (SA) for integrating multi-scale scene-related information to strengthen feature representation. }
	\label{fig:pipeline}
\end{figure*}

\begin{figure*}[t]
	\centering
	\begin{tabular}{c}
		\includegraphics[width=0.98\textwidth]{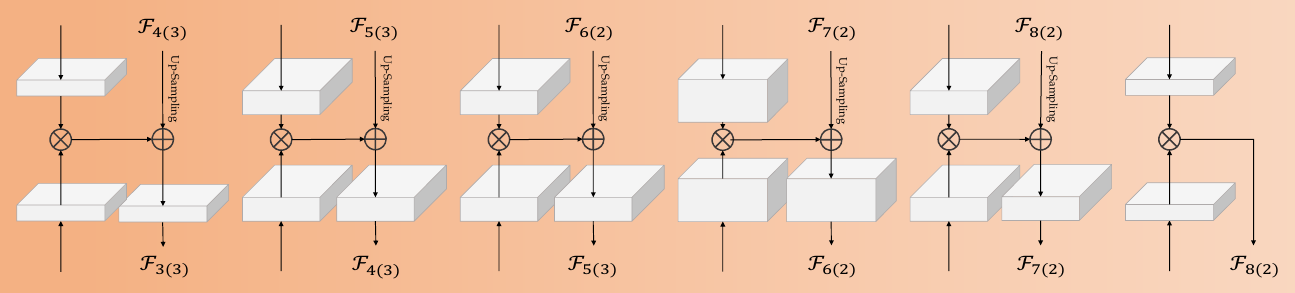}
	\end{tabular}
	\vspace{-0.4cm}
	\caption{The computational flow of our proposed multi-scale feature aggregator.}
	\label{fig:SAM}
\end{figure*}

\section{Proposed Method}
In this section, we present our motivation by analyzing the latent relationship between enhancer and detector. We then construct a scene decomposition module to endow the perception-related ability to the Retinex-based enhancer. Finally, we introduce a semantic aggregation module to strengthen the feature representation. The overall architecture of our proposed method can be found in Figure~\ref{fig:pipeline}. 
\subsection{Two Key Challenges of Enhancer + Detector}

Unlike regular object detection in normal-light environments, the main challenge for low-light object detection is that degraded observations with poor visibility heavily influence feature extraction, causing the accuracy to drop sharply. It is a direct and commonly-used manner to generate new visible data by improving visual quality. It is widely adopted in multiple champion solutions born in the UG2+ Challenge competition (landmark for low-light face detection). Among them, experimental explorations suggest that adopting a classical (e.g., MSRCR~\cite{jobson1997multiscale}) one with poor visual quality is more effective than the latest advanced enhancer. Therefore, the preconceived conclusion ``higher visual quality is more beneficial for detection'' cannot be satisfied. 

Investigating its reason, on the one hand, these two tasks have different objectives, i.e., pixel-level visual-friendly to human eyes (enhancer) and semantic-level perception-friendly to machines (detector). In other words, the directly cascaded pattern aims at improving detection accuracy, and the enhancer is designed for catering to visual quality, rather than detection-desired high-quality data. 
On the other hand, the enhancer indeed ameliorates visual quality but it inevitably destroys the inherent distribution that keeps the same situation as natural images, causing the enhanced data to keep a distinct information discrepancy with regularly captured data (maybe low-light, maybe normal-light), which heavily restricts the feature extraction. Combing the above analyses, we can conclude two key challenges for the paradigm of enhancer + detector, described as
\begin{enumerate}[itemindent=0em]
	\item \textit{What kind of data acquired from the enhancer is required for the detector in low-light scenes?}
	\item \textit{How to narrow down the information discrepancy between enhanced results and regular data?}
\end{enumerate}

To settle these two challenges, in the following, we build a scene decomposition module to enable the Retinex-based enhancer beyond visual quality to generate detection-desired data. Then we construct a semantic aggregation module to fully exploit decomposed scene-related features to reduce the information discrepancy. 


\subsection{Scene Decomposition Module} 
In this part, by rethinking enhancement for detection, we explain the necessity of utilizing the removed illumination. Substantially, we establish a scene decomposition module to acquire the decomposed components. 
\subsubsection{Rethinking Illumination-based Enhancers}
Existing low-light image enhancement techniques~\cite{ma2022toward,liu2021retinex} are mostly developed according to the Retinex theory~\cite{land1971lightness} (formulated as $\mathbf{L}=\mathbf{R}\otimes\mathbf{I}$, where $\otimes$ denotes the element-wise multiplication), this principle describes that low-light observation $\mathbf{L}$ can be decomposed as the normal-light image $\mathbf{R}$ (also called reflectance) and the removed illumination $\mathbf{I}$. We can find a basic fact from this model, i.e., the enhanced result needs to remove the illumination from the low-light observation. This means the enhanced result cannot keep the same information capacity as the original observation and the missing content is exactly the removed illumination based on the information conservation. However, most detectors are established in regular data without information reduction. 

From this perspective, the illumination is abandoned for the enhancer (only needs to focus on the normal-light image), but it should be utilized for the detector to improve the information utilization to the maximum extent. 

\subsubsection{Decomposing Scene for Detection}
As described above, we know the removed illumination is essential for detection. That is to say, the two decomposed components from low-light observations are equally important for detection, rather than only focusing on the visual quality of the single component in the enhancer. Generally, the low-light observations reflect the scene information including objects and background. After performing Retinex theory, these two components still contain the scene information. Therefore, we call the module for generating two decomposed components as Scene Decomposition Module (SDM). 

Here we provide a simple setting for SDM, which consists of three residual blocks and each residual block includes two Conv-BN-ReLU layers.  The 1$\times$1 convolutional layer is used to adjust the number of channels of the feature map at the time of input and output. 
Notably, we would like to emphasize that SDM can be initialized by existing Retinex-based enhancer\footnote{Please refer to Sec.~\ref{sec:discussion} for experimental supports.}.

\begin{table}[t]
	\centering
	\caption{Quantitative results among different versions for illumination-based enhancers including RUAS and SCI.}
	\vspace{-0.3cm}
	\renewcommand\arraystretch{1.3}	
	\setlength{\tabcolsep}{6mm}
	\begin{tabular}{lll}
		\toprule
		\textbf{Method} &\textbf{Description} &\textbf{mAP(\%)}  \\
		\midrule
		RUAS &Fine-tune detector	& 65.98    \\
		RUAS$^{+}$ &Jointly training   &65.70            \\
		RUAS$^{++}$  &T2 (SD = RUAS)           &67.36$_{\color{red}{\uparrow{\textbf{1.66}}}}$   \\
		\midrule
		SCI  &Fine-tune detector  & 65.68 \\
		SCI$^{+}$  &Jointly training   &65.42     \\		
		SCI$^{++}$	  &T2 (SD = SCI)      &68.33$_{\color{red}{\uparrow{\textbf{2.91}}}}$\\	
		\bottomrule
	\end{tabular}
	\vspace{-0.4cm}
	\label{tab:treasure}
\end{table}
%

\subsection{Semantic Aggregation Module}
The previous section has generated two decomposed components by scene decomposition, the next issue is how to exploit them for the detector. Here we build a semantic aggregation module that consists of a weigh-sharing feature extractor and a multi-scale feature aggregator. 
\subsubsection{Weight-Sharing Feature Extractor}\label{sec:overall}
Our defined scene decomposition module aims at generating two decomposed components to perform scene information. Although they are acquired based on the knowledge from low-light image enhancement, their output status should be related to the detection accuracy. Here we adopt an VGG-16 used in DFSD~\cite{li2019dsfd} and S3FD~\cite{zhang2017s3fd} as its basic architecture. We use the weight-sharing backbone network to extract the features of the two decomposed components.

\subsubsection{Multi-Scale Feature Aggregator}
We know that the feature pyramid network~\cite{lin2017feature} is a commonly-used structure for the detector, which can improve the detection accuracy, especially the extremely small objects. 
Through our above-built weight-sharing feature extractor, we can obtain two groups of multi-scale scene-related semantic features in the context space. To effectively integrate them, we define a multi-scale feature aggregator by introducing the Retinex knowledge into the feature pyramid network. This process can be formulated as
\begin{equation}
\left\{
\begin{aligned}	
	\mathcal{F}_{8(2)} &=  \mathcal{F}^{\mathbf{I}}_{8(2)}\otimes\mathcal{F}^{\mathbf{R}}_{8(2)},\\
	\mathcal{F}_{a(b)} &= \mathcal{F}^{\mathbf{I}}_{a(b_{a})}\otimes\mathcal{F}^{\mathbf{R}}_{a(b_{a})}+(\mathcal{F}_{a+1(b_{a+1})})_{\uparrow}, \\
\end{aligned}
\right.
\end{equation}
where $3\leq a\leq 7, b\in\{2,3\}$. $\mathcal{F}^{\mathbf{I}}$ and $\mathcal{F}^{\mathbf{R}}$ represent the generated features from the feature extractor according to illumination and reflectance, respectively. $\mathcal{F}_{a(b)}$ denotes the feature generated in $b$-th convolutional layer in the $a$-th convolution block. We are utilizing six layers of features: $\mathcal{F}_{3(3)}$, $\mathcal{F}_{4(3)}$, $\mathcal{F}_{5(3)}$, $\mathcal{F}_{6(2)}$, $\mathcal{F}_{7(2)}$, and $\mathcal{F}_{8(2)}$. $(\cdot)_{\uparrow}$ represents the upsamling operation. The aggregated operation is defined as the element-wise multiplication $\otimes$. It is because multiplication exactly corresponds to the division used in the scene decomposition. In other words, this way reconstructs the original scene information that existed in the original low-light observation. The detailed computational process can be found in Figure~\ref{fig:SAM}.

\begin{figure*}[t]
	\centering
	\begin{tabular}{c}
		\includegraphics[width=0.985\textwidth]{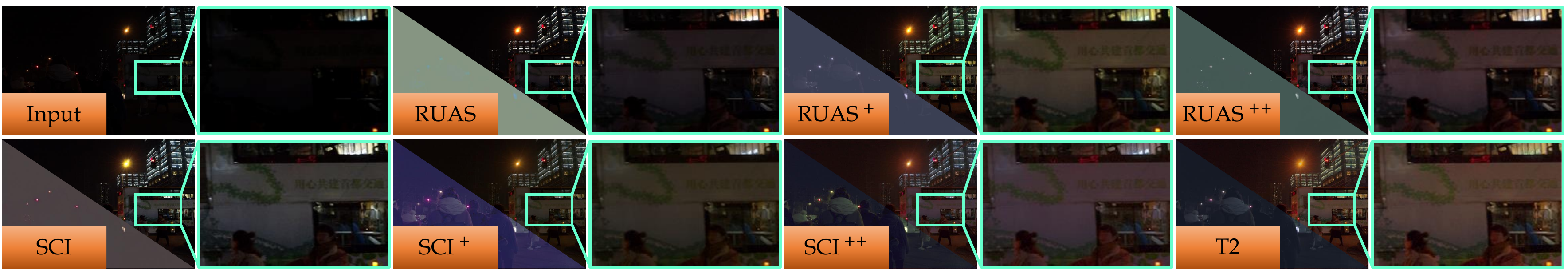}\\
	\end{tabular}
	\vspace{-0.4cm}
	\caption{Comparing decomposed components among different methods on a low-light image. Except the input, the bottom left and top right for each image are illumination and reflectance, respectively. }
	\label{fig:Low-light refl illu}
\end{figure*}

\subsection{Training Loss}
In the training phase, we adopt a multi-task loss function~\cite{liu2016ssd, girshick2015fast} for the overall network. The objective loss function consists of a weighted sum of location loss $\mathcal{L}_{loc}$ and a confidence loss $\mathcal{L}_{conf}$,  formulated as
\begin{equation}
	\mathcal{L}(x,c,l,g)=\frac{1}{N}(\mathcal{L}_{conf}(x,c)+\alpha \mathcal{L}_{loc}(x,l,g)),
\end{equation}
where $\mathcal{L}_{conf}$ is focal loss and $\mathcal{L}_{loc}$ is smooth L1 loss, $\alpha$ represents the trade-off parameter between the two losses. $N$ is the number of the default boxes, ${x}_{ij}= \{ 1, 0\}$ represents an indicator that the $i$-th default box matches the $j$-th ground truth box, ${c}$ represents the predicted class confidence scores and ${l},{g}$ denote the predicted box and the ground truth box, respectively. In this paper, we set $\alpha$ to 1.0. We do not apply any loss constraints specifically to the scene decomposition module.

\subsection{Discussion}\label{sec:discussion}
In this part, we present detailed discussions from two aspects to deeply recognize our proposed method. 

\textit{(1) The illumination is a treasure not trash for the detector}.
Our proposed algorithm is established based on the fact that ``the illumination is a treasure not trash for the detector''. Here we verify this fact from an experimental perspective. 
In Table~\ref{tab:treasure}, we investigate different training patterns for enhancer + detector. Here, we consider two representative enhancers (RUAS and SCI) and the classical SSD detector with Feature Pyramid Networks. 
The fine-tune detector and joint training methods cannot achieve good detection results.
Fortunately, after introducing the illumination for the detector (i.e., RUAS$^{++}$ and SCI$^{++}$), the detection accuracy for these two enhancers all realize a significant boost (see the red-bold texts in the right bottom corner of the third and last rows in Table~\ref{tab:treasure}). Deeply thinking, the performance improvement is benefited from the entire expression for scene information. 
In a word, the experiments can fully verify the necessity of introducing illumination for the detector. Compared with existing manners, we can conclude that ``illumination is a treasure not trash for the detector''.

\textit{(2) Our T2 realizes the detection-oriented enhancement}.
In our designed algorithm, the scene decomposition module is constructed based on physical knowledge (i.e., Retinex theory) for low-light image enhancement. Although we do not define the loss functions related to visual quality in the training phase, this module still implicitly possesses the tendentiousness for enhancing image quality. To verify it, we show the decomposed components among different methods (most methods come from Table~\ref{tab:treasure}) on low-light scenarios, respectively. 

Clearly, as seen in Figure~\ref{fig:Low-light refl illu}, compared with the original enhanced outputs (i.e., RUAS and SCI), the other detection-oriented enhancers, even without visual quality-related constraints, all achieve enhancement effects. Notably, our fully designed T2 exhibits a certain image enhancement effect and is considered to possess friendly-detection features (as observed in the zoomed-in region). Moreover, RUAS$^{++}$ and SCI$^{++}$ perform better visual quality than RUAS$^+$ and SCI$^+$, which also indicates that our T2 also realizes a mutual promotion for visual quality and detection accuracy (see Table~\ref{tab:treasure}).

\begin{figure*}[t]
	\centering
	\begin{tabular}{c@{\extracolsep{0.5em}}c}
		\includegraphics[width=0.48\textwidth]{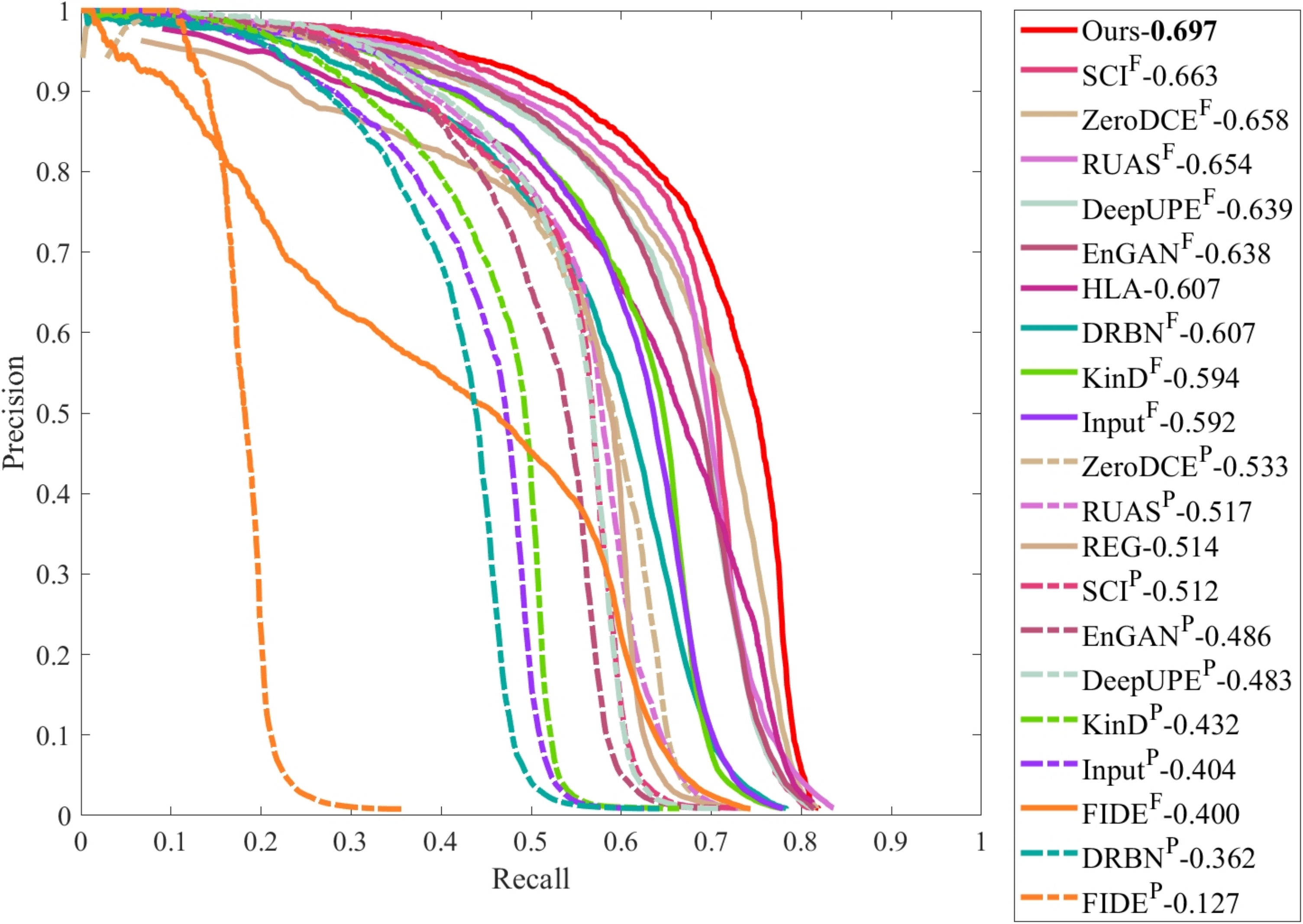}&
		\includegraphics[width=0.48\textwidth]{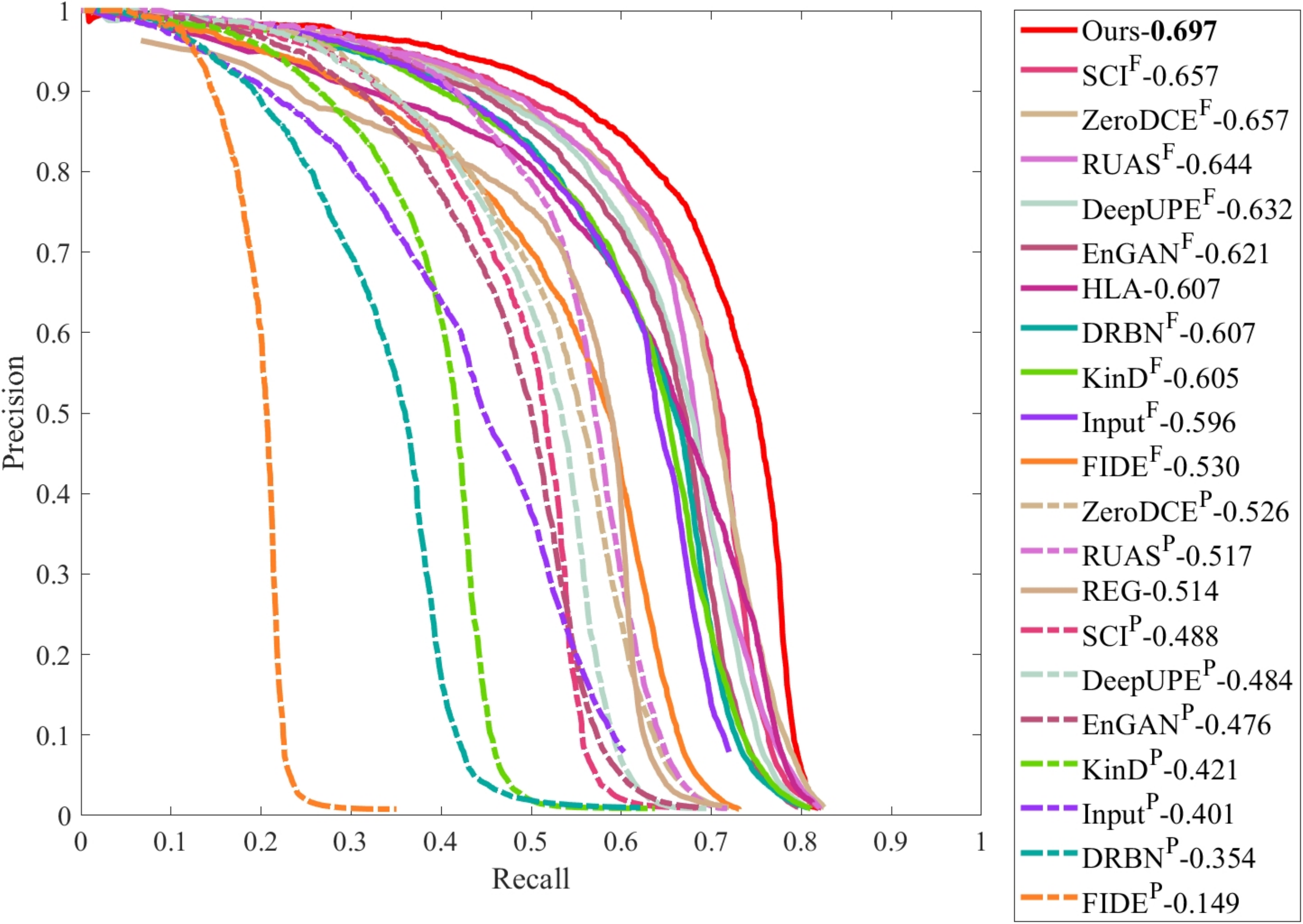} \\
		\footnotesize (a) Pyramidbox &\footnotesize (b) DSFD
	\end{tabular}
	\vspace{-0.4cm}
	\caption{The PR curves of different state-of-the-art methods and our proposed approach on the DARK FACE dataset. $(\cdot)^\mathtt{P}$ and $(\cdot)^\mathtt{F}$ represent the version of using directly pre-trained detectors and fine-tune detectors, respectively.}
	\label{fig:PRcurve}
\end{figure*}
\begin{figure*}[t]
	\centering
	\vspace{-0.2cm}
	\begin{tabular}{c}
		\includegraphics[width=0.985\textwidth]{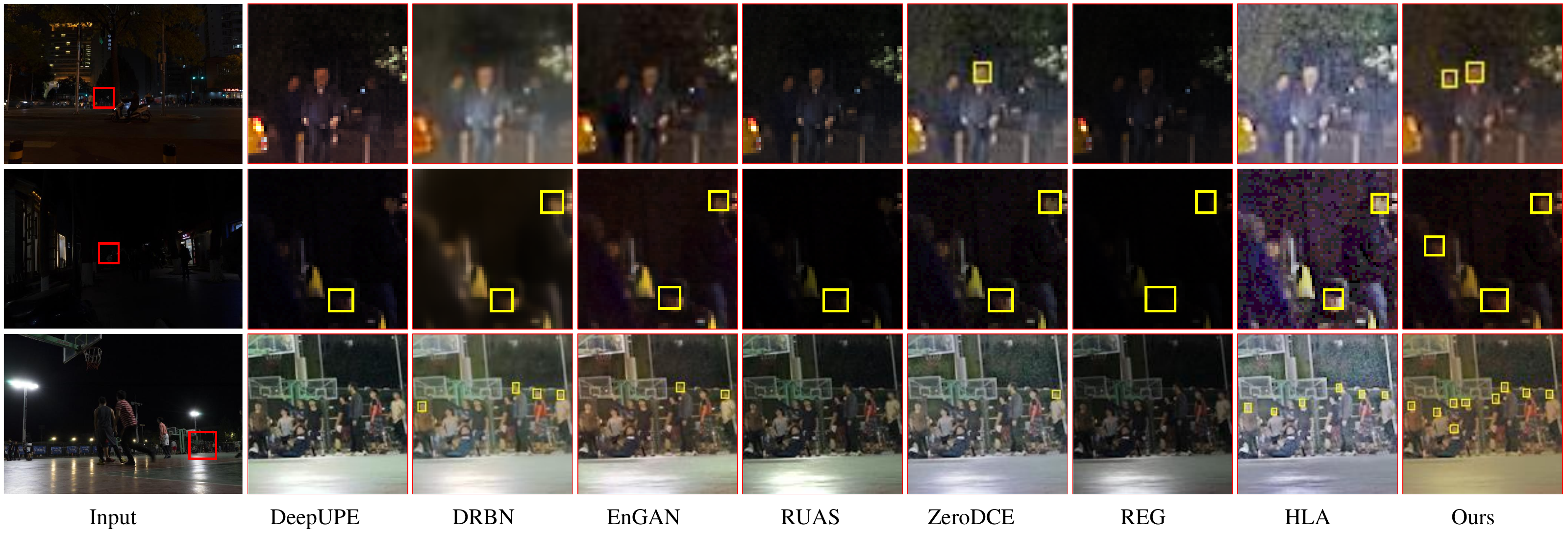}\\
	\end{tabular}
	\vspace{-0.4cm}
	\caption{More visual results of object detection on the DARK FACE dataset. }
	\vspace{-0.2cm}
	\label{fig:DarkFace1}
\end{figure*}

\section{Experimental Results}
\begin{table*}[t]
	\renewcommand\arraystretch{1.2}
	\setlength{\tabcolsep}{4pt}
	\caption{Quantitative results of object detection on the ExDark dataset about the finetuned detector (SSD) on the enhanced results generated by all the compared methods. The best result is in red whereas the second best one is in blue.}
	\vspace{-0.2cm}
	\centering
	\label{table1}
	\begin{tabular}{c|cccccccccccc|c}
		\toprule
		Method &Bicycle &Boat &Bottle &Bus &Car &Cat &Chair &Cup &Dog &Motorbike &People &Table &mAP\\
		\midrule
		Low-Light Input &61.87 &\color{blue}{\textbf{54.42}} &41.77 &85.55 &\color{blue}{\textbf{62.14}} &57.28 &45.81 &39.30 &57.58 &63.68 &54.29 &51.88 &56.30\\
		
		DeepUPE	 &63.99 &47.82 &40.34 &88.92 &62.12 &\color{blue}{\textbf{58.52}} &46.00 &\color{blue}{\textbf{42.06}} &59.07 &\color{red}{\textbf{64.28}} &\color{blue}{\textbf{54.97}} &51.86 &56.66 \\
		
		ZeroDCE	 &\color{red}{\textbf{68.04}} &50.19 &41.61 &\color{blue}{\textbf{88.94}} &60.04 &54.76 &\color{blue}{\textbf{48.62}} &38.25 &58.68 &62.62 &54.79 &\color{blue}{\textbf{55.67}} &\color{blue}{\textbf{56.85}} \\
		
		EnGAN   &60.24 &52.17 &38.31 &\color{red}{\textbf{89.82}} &60.60 &56.49 &47.45 &38.24 &57.75 &63.13 &52.59 &50.04 &55.57 \\
		
		KinD   &62.76 &54.07 &40.36 &84.62 &58.33 &53.07 &42.16 &35.86 &\color{blue}{\textbf{59.90}} &58.34 &53.75 &48.71 &54.33 \\
		
		DRBN   &63.54 &53.14 &36.35 &86.38 &58.83 &51.66 &39.85 &33.73 &55.50 &59.87 &52.12 &50.23 &53.44 \\
		
		RUAS   &63.49 &47.93 &40.61 &85.94 &60.19 &50.14 &44.46 &39.12 &54.37 &60.20 &53.52 &51.91 &54.32 \\
		
		SCI	  &\color{blue}{\textbf{65.41}} &52.08 &\color{red}{\textbf{42.31}} &88.80 &60.44 &53.11 &47.21 &38.29 &56.57 &60.78 &54.80 &49.49 &55.77 \\
		
		Ours  &63.47 &\color{red}{\textbf{57.87}} &\color{blue}{\textbf{41.92}} &88.16 &\color{red}{\textbf{67.09}} &\color{red}{\textbf{64.24}} &\color{red}{\textbf{51.93}} &\color{red}{\textbf{42.85}} &\color{red}{\textbf{60.40}} &\color{blue}{\textbf{63.83}} &\color{red}{\textbf{58.69}} &\color{red}{\textbf{56.48}} &\color{red}{\textbf{59.74}}\\
		\bottomrule
	\end{tabular}
	\label{tab:table_MAP}
\end{table*}
\begin{figure*}[t]
	\centering
	\vspace{-0.2cm}
	\begin{tabular}{c}
		\includegraphics[width=0.985\textwidth]{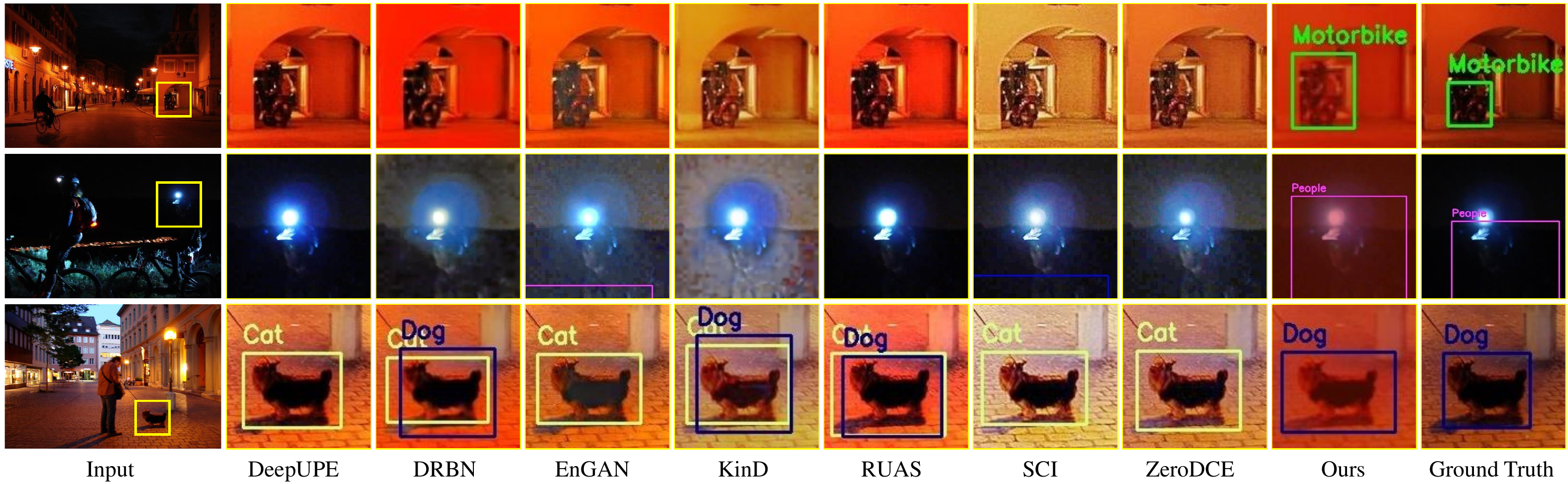}\\
	\end{tabular}
	\vspace{-0.4cm}
	\caption{More visual results of object detection on the ExDark dataset.}
	\label{fig:ExDark}
\end{figure*}

\subsection{Implementation Details}
\subsubsection{Datasets and Metrics} We conducted our experiments using the DARK FACE dataset, which consists of 6000 low-light images captured in real-world environments. The images have a resolution of 1080$\times$720 and contain a variable number of faces, typically ranging from 1 to 20. The labeled faces exhibit a wide range of scales, varying from 1$\times$2 to 335$\times$296. For our experiments, we randomly selected 1000 images for testing, while the remaining images were used for training. To evaluate the performance of our approach, we employed the mean Average Precision (mAP) as the evaluation metric.

\subsubsection{Parameters Setting} For model training, we employed SGD with a momentum of 0.9 and weight decay of 0.0005. The batch size was set to 4, and the initial learning rate was 0.0005. During the training stage, a face anchor was labeled as a positive anchor if it had an Intersection over Union (IoU) of over 0.3 with the ground truth. Furthermore, we maintained a ratio of 3:1 between negative and positive anchors. To ensure consistency, we resized the images to 640$\times$640 for training and 1500$\times$1000 for testing. For effective bounding box selection, we applied non-maximum suppression using Jaccard overlap with a threshold of 0.3, retaining the top 750 high-confidence bounding boxes per image, as inspired by Neubeck and Van Gool's work~\cite{neubeck2006efficient}.

\subsubsection{Compared Methods}
In order to provide a comprehensive evaluation of our detection results, we employed both qualitative and quantitative analyses. Our detection framework is built upon two state-of-the-art face detectors, namely Pyramidbox~\cite{tang2018pyramidbox} and DSFD~\cite{li2019dsfd}. Pyramidbox is a context-assisted single shot face detector that has shown impressive performance in various face detection benchmarks. DSFD is a dual shot face detector that leverages a dual-stage strategy to improve the detection performance, especially for small and hard faces.
To enhance the quality of the input images, we utilized a variety of image enhancement methods including DeepUPE~\cite{wang2019underexposed}, DRBN~\cite{yang2020fidelity}, EnGAN~\cite{jiang2021enlightengan}, FIDE~\cite{xu2020learning}, KinD~\cite{zhang2019kindling}, RUAS~\cite{liu2021retinex}, ZeroDCE~\cite{guo2020zero}, and SCI~\cite{ma2022toward}. In addition to these image enhancement methods, we also compared our results with two dedicated low-light face detection solutions, HLA~\cite{wang2021hla} and REG~\cite{liang2021recurrent}. 


\subsection{Evaluations on the DARK FACE dataset}

\subsubsection{Quantitative comparisons} 
In-depth analysis was conducted by plotting Precision-Recall (PR) curves for two detectors, as illustrated in Figure~\ref{fig:PRcurve}. This comparison revealed two noteworthy conclusions.

Firstly, our proposed method exhibited significant advancements when compared to both the combination of advanced enhancer and detector, as well as mainstream low-light face detectors. To be more specific, our method achieved a remarkable 14.8\% higher mean Average Precision (mAP) than HLA and an impressive 8.2\% higher mAP than the best hybrid scheme, which involved the finetuning version of SCI and DSFD. Importantly, our proposed method consistently outperformed different detectors across various numerical metrics.

Secondly, these findings lend support to our initial motivation that solely relying on visual aesthetics may not necessarily lead to improved detection performance. Although the jointly fine-tuned methods demonstrated some enhancement in performance, there remains substantial room for further improvement.

\subsubsection{Qualitative comparisons} We selected three representative scenarios to demonstrate our detection precision and visual effects straightforwardly, including extreme darkness, dense crowd, and low contrast. Figure~\ref{fig:DarkFace1} demonstrated four groups of visual and detection results comparisons on the DARK FACE dataset. We can observe that our methods obtain two obvious advantages compared with these competitors. Firstly, our method could effectively extract efficient semantic information from illumination for detection and realize high accuracy (\emph{e.g.,} the people in the distance in each group of examples). 
Secondly, benefiting from the guidance of follow-up detection, our method could decompose more suitable illumination to preserve the detection performance. In contrast, our method had less noise interference in detection.

Thus, by leveraging the decomposition and aggregation, the proposed method actually realized the representative feature extraction and utilization, which can simultaneously provide better scene understanding.

\subsection{Evaluations on the ExDARK dataset}
In order to fully verify the performance of detection, we presented more results on the ExDark\cite{loh2019getting} low-light object detection dataset. This dataset involves 12 categories. 737 images were randomly sampled for testing and the remaining 6626 low-light images were used for training and validation. For this dataset, we set the maximum epoch as 100, and the batch size as 32. We used Adam and the learning rate was initialized to 3e-5. We utilized the SSD detection model as the baseline in all methods for the ExDark dataset. We compared our method with the "Enhancer+Detector" detection pattern, which considered low-light image enhancement as a pre-processing method\cite{ma2022pia}. 
Table~\ref{tab:table_MAP} (score-threhold=0.5) reported the detection results on the specific class. It could be easily observed that our method was significantly superior to other methods. Furthermore, Figure~\ref{fig:ExDark} provided the visual comparison among different methods on the ExDark dataset, our method detected more correct objects and made fewer errors.

\begin{table}[t]
	\centering
	\caption{Ablation study among different settings.}
	\vspace{-0.3cm}
	\renewcommand\arraystretch{1.3}
	\setlength{\tabcolsep}{3.5mm}
	\begin{tabular}{cccccc}
		\toprule
		\textbf{Model} &{$\mathbf{L}$}&{$\mathbf{I}$}&$\mathbf{R}$ &\textbf{FPN} &\textbf{mAP (\%)}\\
		\midrule
		A   &\checkmark  &  &  &    &61.13 \\
		B  &\checkmark &  & &\checkmark   &66.42\\
		C  &  & \checkmark & &  &59.28 \\
		D  &  &  & \checkmark &  &66.94\\
		E  & & & \checkmark&\checkmark &67.68\\
		T2    &  & \checkmark & \checkmark &\checkmark &\textbf{69.67} \\
		\bottomrule
	\end{tabular}
	\vspace{-0.4cm}
	\label{tab:analysisComponents}
\end{table}

\subsection{Algorithmic Analyses}
\subsubsection{Ablation Study} We firstly provided five variants (denoted as ``A$\sim$E'') under different settings to demonstrate the proposed mechanisms. Concrete numerical results were reported in Table~\ref{tab:analysisComponents} (when both illumination and reflectance were used as input to the feature extractor, the FPN represented our semantic aggregation FPN). We can summarize three vital observations from this table. Firstly, the illumination layer also maintains enough information capacity, which almost obtains a comparable result compared with the version using low-light inputs $\mathbf{L}$. Furthermore, reflection alone cannot obtain the best performance. Thus, it demonstrates the reasonability to extract the information from illumination. Secondly, 
FPN network plays an important role to improve the performance, which can significantly improve 8.6\% and 0.74 higher mAP when the 
inputs are low-light images $\mathbf{L}$ and reflection $\mathbf{R}$ respectively. Therefore, we use the semantic aggregation FPN in T2. 
For the performance of using scene information including illumination and reflectance, accuracy is significantly improved.


\section{Concluding Remarks}
In this work, we proposed a new low-light object detector consisting of the scene decomposition and the semantic aggregation modules. We first analyzed the latent relationship between enhancer and detector to definitely point out the two key challenges. Then we constructed a scene decomposition module to present scene characteristics. A semantic aggregation composed of a weight-sharing feature extractor and multi-scale feature aggregator is proposed to integrate the scene information in the feature space. Finally, extensive experiments are performed to reveal our superiority. 

\textit{Broader Impacts}. How to effectively characterize scene information is an inherent appeal in designing our method. It is actually a general research focus for a variety of adverse conditions. Decomposing the scenes in the image space and aggregating the scenes in the feature space are crucial measures for our proposed T2. This way provides a new perspective to understand and handle the pattern of enhancer + detector, which will rekindle the enthusiasm to research the task of detection in adverse conditions.

\bibliography{name}

\begin{thebibliography}{47}
\providecommand{\natexlab}[1]{#1}

\bibitem[{Chen et~al.(2018)Chen, Chen, Xu, and Koltun}]{chen2018learning}
Chen, C.; Chen, Q.; Xu, J.; and Koltun, V. 2018.
\newblock Learning to see in the dark.
\newblock In \emph{Proceedings of the IEEE conference on computer vision and
  pattern recognition}, 3291--3300.

\bibitem[{Cui et~al.(2021)Cui, Qi, Gu, You, Zhang, and
  Harada}]{cui2021multitask}
Cui, Z.; Qi, G.-J.; Gu, L.; You, S.; Zhang, Z.; and Harada, T. 2021.
\newblock Multitask aet with orthogonal tangent regularity for dark object
  detection.
\newblock In \emph{Proceedings of the IEEE/CVF International Conference on
  Computer Vision}, 2553--2562.

\bibitem[{Girshick(2015)}]{girshick2015fast}
Girshick, R. 2015.
\newblock Fast r-cnn.
\newblock In \emph{Proceedings of the IEEE international conference on computer
  vision}, 1440--1448.

\bibitem[{Guo et~al.(2020)Guo, Li, Guo, Loy, Hou, Kwong, and
  Cong}]{guo2020zero}
Guo, C.; Li, C.; Guo, J.; Loy, C.~C.; Hou, J.; Kwong, S.; and Cong, R. 2020.
\newblock Zero-reference deep curve estimation for low-light image enhancement.
\newblock In \emph{Proceedings of the IEEE/CVF Conference on Computer Vision
  and Pattern Recognition}, 1780--1789.

\bibitem[{He et~al.(2016)He, Zhang, Ren, and Sun}]{he2016deep}
He, K.; Zhang, X.; Ren, S.; and Sun, J. 2016.
\newblock Deep residual learning for image recognition.
\newblock In \emph{Proceedings of the IEEE conference on computer vision and
  pattern recognition}, 770--778.

\bibitem[{Jiang et~al.(2021)Jiang, Gong, Liu, Cheng, Fang, Shen, Yang, Zhou,
  and Wang}]{jiang2021enlightengan}
Jiang, Y.; Gong, X.; Liu, D.; Cheng, Y.; Fang, C.; Shen, X.; Yang, J.; Zhou,
  P.; and Wang, Z. 2021.
\newblock Enlightengan: Deep light enhancement without paired supervision.
\newblock \emph{IEEE Transactions on Image Processing}, 30: 2340--2349.

\bibitem[{Jin et~al.(2021)Jin, Ma, Liu, and Fan}]{jin2021bridging}
Jin, D.; Ma, L.; Liu, R.; and Fan, X. 2021.
\newblock Bridging the gap between low-light scenes: Bilevel learning for fast
  adaptation.
\newblock In \emph{Proceedings of the 29th ACM International Conference on
  Multimedia}, 2401--2409.

\bibitem[{Jobson, Rahman, and Woodell(1997)}]{jobson1997multiscale}
Jobson, D.~J.; Rahman, Z.-u.; and Woodell, G.~A. 1997.
\newblock A multiscale retinex for bridging the gap between color images and
  the human observation of scenes.
\newblock \emph{IEEE Transactions on Image processing}, 6(7): 965--976.

\bibitem[{Land and McCann(1971)}]{land1971lightness}
Land, E.~H.; and McCann, J.~J. 1971.
\newblock Lightness and retinex theory.
\newblock \emph{Josa}, 61(1): 1--11.

\bibitem[{Li et~al.(2018)Li, Guo, Porikli, and Pang}]{li2018lightennet}
Li, C.; Guo, J.; Porikli, F.; and Pang, Y. 2018.
\newblock LightenNet: A convolutional neural network for weakly illuminated
  image enhancement.
\newblock \emph{Pattern recognition letters}, 104: 15--22.

\bibitem[{Li et~al.(2019)Li, Wang, Wang, Tai, Qian, Yang, Wang, Li, and
  Huang}]{li2019dsfd}
Li, J.; Wang, Y.; Wang, C.; Tai, Y.; Qian, J.; Yang, J.; Wang, C.; Li, J.; and
  Huang, F. 2019.
\newblock DSFD: dual shot face detector.
\newblock In \emph{Proceedings of the IEEE/CVF Conference on Computer Vision
  and Pattern Recognition}, 5060--5069.

\bibitem[{Liang et~al.(2021)Liang, Wang, Quan, Chen, Liu, Ling, and
  Xu}]{liang2021recurrent}
Liang, J.; Wang, J.; Quan, Y.; Chen, T.; Liu, J.; Ling, H.; and Xu, Y. 2021.
\newblock Recurrent exposure generation for low-light face detection.
\newblock \emph{IEEE Transactions on Multimedia}, 24: 1609--1621.

\bibitem[{Lin et~al.(2017)Lin, Doll{\'a}r, Girshick, He, Hariharan, and
  Belongie}]{lin2017feature}
Lin, T.-Y.; Doll{\'a}r, P.; Girshick, R.; He, K.; Hariharan, B.; and Belongie,
  S. 2017.
\newblock Feature pyramid networks for object detection.
\newblock In \emph{Proceedings of the IEEE conference on computer vision and
  pattern recognition}, 2117--2125.

\bibitem[{Liu et~al.(2021{\natexlab{a}})Liu, Fan, Jiang, Liu, and
  Luo}]{liu2021learning}
Liu, J.; Fan, X.; Jiang, J.; Liu, R.; and Luo, Z. 2021{\natexlab{a}}.
\newblock Learning a deep multi-scale feature ensemble and an edge-attention
  guidance for image fusion.
\newblock \emph{IEEE Transactions on Circuits and Systems for Video
  Technology}, 32(1): 105--119.

\bibitem[{Liu et~al.(2021{\natexlab{b}})Liu, Xu, Yang, Fan, and
  Huang}]{liu2021benchmarking}
Liu, J.; Xu, D.; Yang, W.; Fan, M.; and Huang, H. 2021{\natexlab{b}}.
\newblock Benchmarking low-light image enhancement and beyond.
\newblock \emph{International Journal of Computer Vision}, 129: 1153--1184.

\bibitem[{Liu et~al.(2020)Liu, Fan, Zhu, Hou, and Luo}]{liu2020real}
Liu, R.; Fan, X.; Zhu, M.; Hou, M.; and Luo, Z. 2020.
\newblock Real-world underwater enhancement: Challenges, benchmarks, and
  solutions under natural light.
\newblock \emph{IEEE transactions on circuits and systems for video
  technology}, 30(12): 4861--4875.

\bibitem[{Liu et~al.(2012)Liu, Lin, De~la Torre, and Su}]{liu2012fixed}
Liu, R.; Lin, Z.; De~la Torre, F.; and Su, Z. 2012.
\newblock Fixed-rank representation for unsupervised visual learning.
\newblock In \emph{2012 ieee conference on computer vision and pattern
  recognition}, 598--605. IEEE.

\bibitem[{Liu et~al.(2022{\natexlab{a}})Liu, Ma, Ma, Fan, and
  Luo}]{liu2022learning}
Liu, R.; Ma, L.; Ma, T.; Fan, X.; and Luo, Z. 2022{\natexlab{a}}.
\newblock Learning with nested scene modeling and cooperative architecture
  search for low-light vision.
\newblock \emph{IEEE Transactions on Pattern Analysis and Machine
  Intelligence}.

\bibitem[{Liu et~al.(2021{\natexlab{c}})Liu, Ma, Zhang, Fan, and
  Luo}]{liu2021retinex}
Liu, R.; Ma, L.; Zhang, J.; Fan, X.; and Luo, Z. 2021{\natexlab{c}}.
\newblock Retinex-inspired unrolling with cooperative prior architecture search
  for low-light image enhancement.
\newblock In \emph{Proceedings of the IEEE/CVF Conference on Computer Vision
  and Pattern Recognition}, 10561--10570.

\bibitem[{Liu et~al.(2016)Liu, Anguelov, Erhan, Szegedy, Reed, Fu, and
  Berg}]{liu2016ssd}
Liu, W.; Anguelov, D.; Erhan, D.; Szegedy, C.; Reed, S.; Fu, C.-Y.; and Berg,
  A.~C. 2016.
\newblock Ssd: Single shot multibox detector.
\newblock In \emph{European conference on computer vision}, 21--37. Springer.

\bibitem[{Liu et~al.(2022{\natexlab{b}})Liu, Ren, Yu, Guo, Zhu, and
  Zhang}]{liu2022image}
Liu, W.; Ren, G.; Yu, R.; Guo, S.; Zhu, J.; and Zhang, L. 2022{\natexlab{b}}.
\newblock Image-adaptive YOLO for object detection in adverse weather
  conditions.
\newblock In \emph{Proceedings of the AAAI Conference on Artificial
  Intelligence}, volume~36, 1792--1800.

\bibitem[{Liu et~al.(2021{\natexlab{d}})Liu, Lin, Cao, Hu, Wei, Zhang, Lin, and
  Guo}]{liu2021swin}
Liu, Z.; Lin, Y.; Cao, Y.; Hu, H.; Wei, Y.; Zhang, Z.; Lin, S.; and Guo, B.
  2021{\natexlab{d}}.
\newblock Swin transformer: Hierarchical vision transformer using shifted
  windows.
\newblock In \emph{Proceedings of the IEEE/CVF international conference on
  computer vision}, 10012--10022.

\bibitem[{Loh and Chan(2019)}]{loh2019getting}
Loh, Y.~P.; and Chan, C.~S. 2019.
\newblock Getting to know low-light images with the exclusively dark dataset.
\newblock \emph{Computer Vision and Image Understanding}, 178: 30--42.

\bibitem[{Lore, Akintayo, and Sarkar(2017)}]{lore2017llnet}
Lore, K.~G.; Akintayo, A.; and Sarkar, S. 2017.
\newblock LLNet: A deep autoencoder approach to natural low-light image
  enhancement.
\newblock \emph{Pattern Recognition}, 61: 650--662.

\bibitem[{Lv, Li, and Lu(2021)}]{lv2021attention}
Lv, F.; Li, Y.; and Lu, F. 2021.
\newblock Attention guided low-light image enhancement with a large scale
  low-light simulation dataset.
\newblock \emph{International Journal of Computer Vision}, 129(7): 2175--2193.

\bibitem[{Ma et~al.(2023)Ma, Jin, An, Liu, Fan, and Liu}]{ma2023bilevel}
Ma, L.; Jin, D.; An, N.; Liu, J.; Fan, X.; and Liu, R. 2023.
\newblock Bilevel Fast Scene Adaptation for Low-Light Image Enhancement.
\newblock \emph{arXiv preprint arXiv:2306.01343}.

\bibitem[{Ma et~al.(2022{\natexlab{a}})Ma, Ma, Liu, Fan, and
  Luo}]{ma2022toward}
Ma, L.; Ma, T.; Liu, R.; Fan, X.; and Luo, Z. 2022{\natexlab{a}}.
\newblock Toward Fast, Flexible, and Robust Low-Light Image Enhancement.
\newblock In \emph{Proceedings of the IEEE/CVF Conference on Computer Vision
  and Pattern Recognition}, 5637--5646.

\bibitem[{Ma et~al.(2022{\natexlab{b}})Ma, Ma, Fan, Luo, and Liu}]{ma2022pia}
Ma, T.; Ma, L.; Fan, X.; Luo, Z.; and Liu, R. 2022{\natexlab{b}}.
\newblock PIA: Parallel Architecture with Illumination Allocator for Joint
  Enhancement and Detection in Low-Light.
\newblock In \emph{Proceedings of the 30th ACM International Conference on
  Multimedia}, 2070--2078.

\bibitem[{Neubeck and Van~Gool(2006)}]{neubeck2006efficient}
Neubeck, A.; and Van~Gool, L. 2006.
\newblock Efficient non-maximum suppression.
\newblock In \emph{18th International Conference on Pattern Recognition
  (ICPR'06)}, volume~3, 850--855. IEEE.

\bibitem[{Piao et~al.(2019)Piao, Ji, Li, Zhang, and Lu}]{piao2019depth}
Piao, Y.; Ji, W.; Li, J.; Zhang, M.; and Lu, H. 2019.
\newblock Depth-induced multi-scale recurrent attention network for saliency
  detection.
\newblock In \emph{Proceedings of the IEEE/CVF international conference on
  computer vision}, 7254--7263.

\bibitem[{Piao et~al.(2020)Piao, Rong, Zhang, Ren, and Lu}]{piao2020a2dele}
Piao, Y.; Rong, Z.; Zhang, M.; Ren, W.; and Lu, H. 2020.
\newblock A2dele: Adaptive and attentive depth distiller for efficient RGB-D
  salient object detection.
\newblock In \emph{Proceedings of the IEEE/CVF conference on computer vision
  and pattern recognition}, 9060--9069.

\bibitem[{Tang et~al.(2020)Tang, Li, Peng, and Tang}]{tang2020blockmix}
Tang, H.; Li, Z.; Peng, Z.; and Tang, J. 2020.
\newblock Blockmix: meta regularization and self-calibrated inference for
  metric-based meta-learning.
\newblock In \emph{Proceedings of the 28th ACM international conference on
  multimedia}, 610--618.

\bibitem[{Tang et~al.(2022)Tang, Yuan, Li, and Tang}]{tang2022learning}
Tang, H.; Yuan, C.; Li, Z.; and Tang, J. 2022.
\newblock Learning attention-guided pyramidal features for few-shot
  fine-grained recognition.
\newblock \emph{Pattern Recognition}, 130: 108792.

\bibitem[{Tang et~al.(2018)Tang, Du, He, and Liu}]{tang2018pyramidbox}
Tang, X.; Du, D.~K.; He, Z.; and Liu, J. 2018.
\newblock Pyramidbox: A context-assisted single shot face detector.
\newblock In \emph{Proceedings of the European conference on computer vision
  (ECCV)}, 797--813.

\bibitem[{Wang et~al.(2019)Wang, Zhang, Fu, Shen, Zheng, and
  Jia}]{wang2019underexposed}
Wang, R.; Zhang, Q.; Fu, C.-W.; Shen, X.; Zheng, W.-S.; and Jia, J. 2019.
\newblock Underexposed photo enhancement using deep illumination estimation.
\newblock In \emph{Proceedings of the IEEE/CVF Conference on Computer Vision
  and Pattern Recognition}, 6849--6857.

\bibitem[{Wang et~al.(2022)Wang, Wang, Yang, and Liu}]{wang2022unsupervised}
Wang, W.; Wang, X.; Yang, W.; and Liu, J. 2022.
\newblock Unsupervised face detection in the dark.
\newblock \emph{IEEE Transactions on Pattern Analysis and Machine
  Intelligence}, 45(1): 1250--1266.

\bibitem[{Wang, Yang, and Liu(2021)}]{wang2021hla}
Wang, W.; Yang, W.; and Liu, J. 2021.
\newblock Hla-face: Joint high-low adaptation for low light face detection.
\newblock In \emph{Proceedings of the IEEE/CVF Conference on Computer Vision
  and Pattern Recognition}, 16195--16204.

\bibitem[{Wei et~al.(2018)Wei, Wang, Yang, and Liu}]{wei2018deep}
Wei, C.; Wang, W.; Yang, W.; and Liu, J. 2018.
\newblock Deep retinex decomposition for low-light enhancement.
\newblock \emph{arXiv preprint arXiv:1808.04560}.

\bibitem[{Wu, Lin, and Zha(2019)}]{wu2019essential}
Wu, J.; Lin, Z.; and Zha, H. 2019.
\newblock Essential tensor learning for multi-view spectral clustering.
\newblock \emph{IEEE Transactions on Image Processing}, 28(12): 5910--5922.

\bibitem[{Xu et~al.(2020)Xu, Yang, Yin, and Lau}]{xu2020learning}
Xu, K.; Yang, X.; Yin, B.; and Lau, R.~W. 2020.
\newblock Learning to restore low-light images via
  decomposition-and-enhancement.
\newblock In \emph{Proceedings of the IEEE/CVF Conference on Computer Vision
  and Pattern Recognition}, 2281--2290.

\bibitem[{Xue et~al.(2022)Xue, He, Ma, Wang, Fan, and Liu}]{xue2022best}
Xue, X.; He, J.; Ma, L.; Wang, Y.; Fan, X.; and Liu, R. 2022.
\newblock Best of Both Worlds: See and Understand Clearly in the Dark.
\newblock In \emph{Proceedings of the 30th ACM International Conference on
  Multimedia}, 2154--2162.

\bibitem[{Yang et~al.(2020{\natexlab{a}})Yang, Wang, Fang, Wang, and
  Liu}]{yang2020fidelity}
Yang, W.; Wang, S.; Fang, Y.; Wang, Y.; and Liu, J. 2020{\natexlab{a}}.
\newblock From fidelity to perceptual quality: A semi-supervised approach for
  low-light image enhancement.
\newblock In \emph{Proceedings of the IEEE/CVF conference on computer vision
  and pattern recognition}, 3063--3072.

\bibitem[{Yang et~al.(2020{\natexlab{b}})Yang, Yuan, Ren, Liu, Scheirer, Wang,
  Zhang, Zhong, Xie, Pu et~al.}]{yang2020advancing}
Yang, W.; Yuan, Y.; Ren, W.; Liu, J.; Scheirer, W.~J.; Wang, Z.; Zhang, T.;
  Zhong, Q.; Xie, D.; Pu, S.; et~al. 2020{\natexlab{b}}.
\newblock Advancing image understanding in poor visibility environments: A
  collective benchmark study.
\newblock \emph{IEEE Transactions on Image Processing}, 29: 5737--5752.

\bibitem[{Yuan et~al.(2019)Yuan, Yang, Ren, Liu, Scheirer, and
  Wang}]{yuan2019ug}
Yuan, Y.; Yang, W.; Ren, W.; Liu, J.; Scheirer, W.~J.; and Wang, Z. 2019.
\newblock UG2+Track 2: A Collective Benchmark Effort for Evaluating and
  Advancing Image Understanding in Poor Visibility Environments.
\newblock \emph{arXiv preprint arXiv:1904.04474}.

\bibitem[{Zhang et~al.(2020)Zhang, Ren, Piao, Rong, and Lu}]{zhang2020select}
Zhang, M.; Ren, W.; Piao, Y.; Rong, Z.; and Lu, H. 2020.
\newblock Select, supplement and focus for RGB-D saliency detection.
\newblock In \emph{Proceedings of the IEEE/CVF conference on computer vision
  and pattern recognition}, 3472--3481.

\bibitem[{Zhang et~al.(2017)Zhang, Zhu, Lei, Shi, Wang, and Li}]{zhang2017s3fd}
Zhang, S.; Zhu, X.; Lei, Z.; Shi, H.; Wang, X.; and Li, S.~Z. 2017.
\newblock S3fd: Single shot scale-invariant face detector.
\newblock In \emph{Proceedings of the IEEE international conference on computer
  vision}, 192--201.

\bibitem[{Zhang, Zhang, and Guo(2019)}]{zhang2019kindling}
Zhang, Y.; Zhang, J.; and Guo, X. 2019.
\newblock Kindling the darkness: A practical low-light image enhancer.
\newblock In \emph{Proceedings of the 27th ACM international conference on
  multimedia}, 1632--1640.

\end{thebibliography}

\end{document}